\documentclass[twoside,11pt]{article}

\usepackage{jmlr2e}
\pdfoutput=1

\usepackage{graphicx} 
\usepackage{subfigure}
\usepackage{natbib}
\usepackage{wrapfig}

\usepackage{algorithm}
\usepackage{algorithmic}

\usepackage{hyperref}

\ShortHeadings{Distance Metric Learning for Kernel Machines}{}
\firstpageno{1}

\begin{document}

\title{Distance Metric Learning for Kernel Machines}

\author{\name Zhixiang (Eddie) Xu \email xuzx@cse.wustl.edu \\
       \addr Department of Computer Science and Engineering\\
       Washington University in St. Louis\\
       Saint Louis, MO 63130, USA
       \AND
       \name Kilian Q.\ Weinberger \email kilian@wustl.edu \\
       \addr Department of Computer Science and Engineering\\
       Washington University in St. Louis\\
       Saint Louis, MO 63130, USA
       \AND
       \name Olivier Chapelle \email Olivier@chapelle.cc \\
       \addr Criteo \\
	   Palo Alto, CA 94301 }

\editor{}

\maketitle

\newcommand{\argmax}{\operatornamewithlimits{argmax}}
\newcommand{\argmin}{\operatornamewithlimits{argmin}}
\newcommand{\x}{\mathbf{x}}
\newcommand{\ab}{\mathbf{\alpha}}
\newcommand{\y}{\mathbf{y}}
\newcommand{\Lb}{\mathbf{L}}
\newcommand{\Mb}{\mathbf{M}}
\newcommand{\Ib}{\mathbf{I}}
\newcommand{\Kb}{\mathbf{K}}
\newcommand{\Hb}{\mathbf{H}}
\newcommand{\fix}{\marginpar{FIX}}
\newcommand{\new}{\marginpar{NEW}}

\begin{abstract}
\noindent Recent work in metric learning has significantly improved the state-of-the-art in $k$-nearest neighbor classification. 
Support vector machines (SVM), particularly with RBF kernels, are amongst the most popular  classification algorithms that uses distance metrics to compare examples. This paper provides an empirical analysis of the efficacy of three of the most popular Mahalanobis metric learning algorithms as pre-processing for SVM training. We show that none of these algorithms generate metrics that lead to particularly satisfying improvements for SVM-RBF classification. 
As a remedy we introduce support vector metric learning (SVML), a novel algorithm that seamlessly combines the learning of a Mahalanobis metric with the training of the RBF-SVM parameters. We demonstrate the capabilities of SVML on nine benchmark data sets of varying sizes and difficulties. In our study, SVML  
outperforms all alternative state-of-the-art metric learning algorithms in terms of accuracy and establishes itself as a serious alternative to the standard Euclidean metric with model selection by cross validation. 
\end{abstract}

{}

\section{Introduction}
\label{sec_introduction}
Many machine learning algorithms, such as $k$-nearest neighbors (kNN)~\citep{Cover67}, $k$-means~\citep{lloid1982least} or support vector machines (SVM)~\citep{cortes1995support} with shift-invariant kernels, require a distance metric to compare instances. These algorithms rely on the assumption that semantically similar inputs are close, whereas semantically dissimilar inputs are far away. Traditionally, the most commonly used distance metrics are uninformed norms, like the Euclidean distance. In many cases, such uninformed norms are sub-optimal. To illustrate this point, imagine a scenario where two researchers want to classify the same data set of facial images. The first one classifies people by age, the second by gender. Clearly, two images that are similar according to the first researcher's setting might be dissimilar according to the second's. 

Uninformed norms ignore two important contextual components of most machine learning applications. First, in supervised learning the data is accompanied by labels which essentially encode the semantic definition of similarity. Second, the user knows which machine learning algorithm will be used. Ideally, the distance metric should be tailored to the particular setting at hand, incorporating both of these considerations. 


A generalization of the Euclidean distance is the Mahalanobis distance~\citep{mahalanobis1936generalized}. Recent years have witnessed a surge of innovation on  Mahalanobis pseudo-metric  learning~\citep{Davis07itml,Globerson06,Goldberger05,Shental02,Weinberger06distancemetric}. Although these algorithms use different methodologies, the common theme is moving similar inputs closer and dissimilar inputs further away --- where similarity is generally defined through class membership. This transformation can be learned through convex optimization with pairwise constraints~\citep{Davis07itml, Weinberger06distancemetric}, gradient descent with soft neighborhood assignments~\citep{Goldberger05}, or spectral methods based on second-order statistics~\citep{Shental02}. 

Typically, the Mahalanobis metric learning algorithms are used in a two-step approach. First the metric is learned, then it is used for training the classifier or clustering algorithm of choice. 
 The resulting distances are semantically more meaningful than the plain Euclidean distance as they reflect the label information. This makes them  particularly suited for the $k$-nearest neighbor rule, leading to large improvements in classification error~\citep{Davis07itml,Globerson06,Goldberger05,Shental02,Weinberger06distancemetric}. In fact, several algorithms explicitly mimic the $k$-NN rule and minimize a surrogate loss function of the corresponding leave-one-out classification error on the training set~\citep{Goldberger05,Weinberger06distancemetric}.  

Although the $k$-nearest neighbor rule can be a powerful classifier especially in settings with many classes, it comes with certain limitations. For example, the entire training data needs to be stored and processed during test time. Also, in settings with fewer classes (especially binary) it is generally outperformed by Support Vector Machines~\citep{cortes1995support}. 
%
Because of their high reliability as out-of-the-box classifiers, SVMs have become one of the quintessential classification algorithms in many areas of science and beyond. An important part of using SVMs is the right choice of kernel. The kernel function $k(\x_i,\x_j)$ encodes the similarity between two input vectors $\x_i$ and $\x_j$. There are many possible choices for such a  kernel function. One of the most commonly used kernels is the Radial Basis Function (RBF) kernel~\citep{Scholkopf02}, which itself relies on a distance metric. 

This paper considers metric learning for support vector machines. As a first contribution, we review and investigate several recently published kNN metric learning algorithms for the use of SVMs with RBF kernels. 
We demonstrate empirically that these approaches do not reliably improve SVM classification results up to statistical significance. 
As a second contribution, we derive a novel metric learning algorithm that specifically incorporates the SVM loss function during training. 
Here, we learn the metric to minimize the validation error of the SVM prediction at the same time that we train the SVM. This is in contrast to the two-step approach of first learning a metric and then training the SVM classifier with the resulting kernel.  This algorithm, which we refer to as Support Vector Metric Learning (SVML), is particularly useful for three reasons. First, it achieves state-of-the-art classification results and clearly outperforms other metric learning algorithms that are not explicitly geared towards SVM classification. Second, it provides researchers outside of the machine-learning community a convenient way to automatically pre-process their data before applying SVMs.


This paper is organized as follows. In Section~\ref{sec_svm}, we introduce necessary notation and review some background on SVMs. In Section~\ref{sec_metric} we introduce several recently published metric learning algorithms and report results for SVM-RBF classification. In Section~\ref{sec_method} we derive the SVML algorithm and some interesting variations. In Section~\ref{sec_results}, we evaluate SVML on nine publicly available data sets featuring a multitude of different data types and learning tasks. We discuss related work in Section~\ref{sec_related} and conclude in Section~\ref{sec_conclusion}.

\section{Support Vector Machines}
\label{sec_svm}
Let the training data consist of input vectors $\left\{\x_1,\dots,\x_n\right\}\in{\cal R}^d$ with corresponding discrete class labels $\left\{y_1,\dots,y_n\right\}\in\{+1,-1\}$. Although our framework can easily be applied in a multi-class setting, for the sake of simplicity we focus on binary scenarios, restricting $y_i$ to two classes. 

There are several reasons why SVMs are particularly popular classifiers.  First, they are linear classifiers that involve a quadratic minimization problem, which is convex and guarantees perfect reproducibility. Furthermore, the maximum margin philosophy leads to reliably good generalization error rates~\citep{Vapnik98}. But perhaps most importantly, the \emph{kernel-trick}~\citep{Scholkopf02} allows SVMs to generate highly non-linear decision boundaries with low computational overhead. 
More explicitly, the kernel-trick maps the input vectors $\x_i$ implicitly into a higher (possibly infinite) dimensional feature space with a non-linear transformation $\phi:{\cal R}^d\rightarrow {\cal H}$. Training a linear classifier directly in this high dimensional feature space ${\cal H}$ would be computationally infeasible if the vectors $\phi(\x_i)$ were accessed explicitly. However, SVMs can be trained completely in terms of inner-products between input vectors. With careful selection of $\phi()$, the inner-product $\phi(\x_i)^\top\phi(\x_j)$ can be computed efficiently even if computation of the mapping $\phi()$ itself is infeasible. Let the kernel function be $k(\x_i,\x_j)=\phi(\x_i)^\top\phi(\x_j)$ and the $n\times n$ kernel matrix be $\Kb_{ij}=k(\x_i,\x_j)$. The optimization problem of SVM training can be expressed entirely in terms of the kernel matrix $\Kb$.  For the sake of brevity, we omit the derivation and refer the interested reader to one of many detailed descriptions thereof~\citep{Scholkopf02}. The resulting classification rule of a test point $\x_t$ becomes
\begin{equation}
	h(\x_t)=\textrm{sign}(\sum_{j=1}^{n}\alpha_j y_j k(\x_j,\x_t)+b),\label{eq:svm_h}
\end{equation}
where $b$ is the offset of the separating hyperplane and $\alpha_1,\dots,\alpha_n$ are the dual variables corresponding to the inputs $\x_1,\dots,\x_n$. In the case of the hard-margin SVM, the parameters $\alpha_i$ are learned with the following quadratic optimization problem 
\begin{eqnarray}
	&& \min_{\alpha_1,\dots,\alpha_n}\sum_{i=1}^n \alpha_i - \frac{1}{2} \displaystyle\sum_{i,j=1}^n \alpha_i \alpha_j y_i y_j 	K(x_i,x_j) \nonumber \\
	&& \textrm{ subject to : } \displaystyle\sum_{i=1}^n \alpha_i y_i = 0 \textrm{ and }  \alpha_i \geq 0.\label{eq:svm_hard}
\end{eqnarray}
The optimization problem~(\ref{eq:svm_hard}) ensures that all inputs $\x_i$ with label $y_i=-1$ are on one side of the hyperplane, and those with label $y_j=+1$ are on the other. These hard constraints might not always be feasible, or in the interest of minimizing the generalization error (\emph{e.g.} in the case of noisy data). Relaxing the constraints can be performed simply by altering the kernel matrix to
\begin{equation}
	\Kb\leftarrow \Kb+\frac{1}{C}\Ib^{n\times n}.\label{eq:ridge}
\end{equation}
Solving~(\ref{eq:svm_hard}) with a kernel matrix~(\ref{eq:ridge}) is equivalent to a squared-penalty of the violations of the separating hyperplane~\citep{cortes1995support}. This formulation requires no explicit slack variables in the optimization problem and therefore simplifies the derivations of the following sections.

\subsection{RBF Kernel}
There are many different kernel functions that are suitable for SVMs. In fact, any function $k(\cdot,\cdot)$ is a well-defined kernel as long as it is positive semi-definite~\citep{Scholkopf02}. The Radial Basis Function (RBF)-Kernel is defined as follows:
\begin{equation}
k(\x_i,\x_j) =e^{-d^2(\x_i,\x_j)},\label{eq:rbf}
\end{equation}
where $d(\cdot,\cdot)$ is a dissimilarity measure that must ensure positive semidefiniteness of $k(\cdot,\cdot)$. The most common choice is the re-scaled squared Euclidean distance, defined as
\begin{equation}
	d^2(\x_i,\x_j)=\frac{1}{\sigma^2}(\x_i-\x_j)^\top(\x_i-\x_j),
\end{equation}
with \emph{kernel width} $\sigma\!>\!0$.
The RBF-kernel is one of the most popular kernels and yields reliable good classification results. Also, with careful selection of $C$, SVMs with RBF-kernels have been shown to be consistent classifiers~\citep{Steinwart2001}.

\subsection{Relationship with kNN}
The $k$-nearest neighbor classification rule predicts the label of a test point $\x_t$ through a majority vote amongst its $k$ nearest neighbors. Let $\eta_{j}(\x_t)\in\{0,1\}$ be the neighborhood indicator function of a test point $\x_t$, where $\eta_j(\x_t)=1$ if and only if $\x_j$ is one of the $k$ nearest neighbors of $\x_t$. The kNN classification rule can then be expressed as 
\begin{equation}
	h(\x_t)=\textrm{sign}(\sum_{j=1}^n\eta_j(\x_t) y_j ).\label{eq:knn_rule}
\end{equation}
Superficially, the classification rule in~(\ref{eq:knn_rule}) very much resembles~(\ref{eq:svm_h}). In fact, one can interpret the SVM-RBF classification rule in (\ref{eq:svm_h}) as a \emph{``soft''}-nearest neighbor rule. Instead of the zero-one step function $\eta_j(\x_t)$, the training points are weighted  by $\alpha_jk(\x_t,\x_j)$. The classification is still local-neighborhood based, as  $k(\x_t,\x_j)$ decreases exponentially  with increasing distance $d(\x_t,\x_j)$. The SVM optimization in~(\ref{eq:svm_hard}) assigns appropriate weights $\alpha_j\geq 0$ to ensure that, on the leave-one-out training set, the majority vote is correct for all data points by a large margin.

\section{Metric Learning}
\label{sec_metric}
It is natural to ask if the SVM classification rule can be improved with better adjusted metrics than the Euclidean distance. 
A commonly used generalization of the Euclidean metric is the Mahalanobis metric~\citep{mahalanobis1936generalized}, defined as 
\begin{equation}
	d_{\Mb}(\x_i,\x_j)=\sqrt{(\x_i-\x_j)^\top \Mb (\x_i-\x_j)},\label{eq:mahala}
\end{equation}
for some matrix $\Mb\in{\cal R}^{d\times d}$. The matrix $\Mb$ must be semi-positive definite ($\Mb\succeq 0$), which is equivalent to requiring that it can be decomposed into $\Mb=\Lb^\top\Lb$, for some matrix $\Lb\in{\cal R}^{r\times d}$. 
If $\Mb=\Ib^{d\times d}$, where $\Ib^{d\times d}$ refers to the identity matrix in ${\cal R}^{d\times d}$,  (\ref{eq:mahala}) reduces to the Euclidean metric. Otherwise, it is equivalent to the Euclidean distance after the transformation $\x_i\rightarrow \Lb\x_i$. 
Technically, if $\Mb=\Lb^\top \Lb$ is a singular matrix, the corresponding Mahalanobis distance is a  \emph{pseudo-metric}\footnote{A pseudo-metric is not require to preserve identity, i.e. $d(\x_i,\x_j)=0\ \iff \x_i=\x_j$. }. Because the distinction between pseudo-metric and metric is unimportant for this work, we refer to both as \emph{metrics}. As the distance in~(\ref{eq:mahala}) can equally be parameterized by $\Lb$ and $\Mb$ we use $d_{\Mb}$ and $d_{\Lb}$ interchangeably. 

In the following section, we will introduce several approaches that focus on Mahalanobis metric learning for $k$-nearest neighbor classification.\\

\subsection{Neighborhood component analysis}
\cite{Goldberger05} propose 
Neighborhood Component Analysis (NCA), which minimizes the expected leave-one-out classification error under a probabilistic neighborhood assignment. 
For each data point or query, the neighbors are drawn from a softmax probability distribution. The probability of sampling $\x_j$ as a neighbor of $\x_i$ is given by:
\begin{equation}
	p_{ij} = \left \{
	\begin{array}{c c}
		\frac{e^{-d^2_{\Lb}(\x_i,\x_j)}}{\sum_{k\neq i}e^{-d^2_{\Lb}(\x_i,\x_k)}} & \textrm{if}\hspace{4pt} i\neq j \\
		0 & \textrm{if} \hspace{4pt} i=j \\
	\end{array}	\right.	~\label{eq:sample}
\end{equation}
Let us define an indicator variable $y_{ij}\in\{0,1\}$ where $y_{ij}=1$ if and only if $y_i=y_j$. 
With the probability assignment described in (\ref{eq:sample}), we can easily compute the expectation of the leave-one-out classification \emph{accuracy} as
\begin{equation}
	A_{loo} = \frac{1}{n}\displaystyle \sum_{i=1}^n \sum_{j=1}^n p_{ij}y_{ij}. \label{eq:expectnca}
\end{equation} 
NCA uses gradient ascent to maximize~(\ref{eq:expectnca}). 
The advantage of the probabilistic framework over regular kNN is that (\ref{eq:expectnca}) is a continuous, differentiable function with respect to the linear transformation $\Lb$. 
By contrast, the leave-one-out error of regular kNN is not continuous or differentiable. 
The two down-sides of NCA are its relatively high computational complexity and non-convexity of the objective. 

\subsection{Large Margin Nearest Neighbor Classification}
Large Margin Nearest Neighbor (LMNN), proposed by~\cite{Weinberger06distancemetric}, also mimics the leave-one-out error of kNN. Unlike NCA, LMNN employs a convex loss function, and encourages local neighborhoods to have the same labels by pushing data points with different labels away and pulling those with similar labels closer. The authors introduce the concept of \emph{target neighbors}. A target neighbor of a training datum $\x_i$ are data points in the training set that \emph{should ideally be} the nearest neighbors (e.g. the closest points under the Euclidean metric with the same class label). LMNN moves these points closer by minimizing 
\begin{equation}
	\sum_{j\leadsto i} d_{\Mb}(\x_i,\x_j),\label{eq:lmnnobj}
\end{equation}
where $j\leadsto i$ indicates that $\x_j$ is a target neighbor of $\x_i$. 
In addition to the objective~(\ref{eq:lmnnobj}), LMNN also enforces that no datum with a different label can be closer than a target neighbor. In particular, let $\x_i$ be a training point and $\x_j$ one of its target neighbors. Any point $\x_k$ of \emph{different class membership} than $\x_i$ should be further away than $\x_j$ by a large margin. LMNN encodes this relationship as linear constraints with respect to $\Mb$. 
\begin{equation}
d^2_{\Mb}(\x_i,\x_k)  \ge d^2_{\Mb}(\x_i,\x_j)+1 \label{eq:largemargin}
\end{equation}
%
LMNN uses semidefinite programming to minimize~(\ref{eq:lmnnobj}) with respect to~(\ref{eq:largemargin}). 
To account for the natural limitations of a single linear transformation the authors introduce slack variables. More explicitly, for each triple  $(i,j,k)$, where $\x_j$ is a target neighbor of $\x_i$ and $y_k\neq y_i$, they introduce $\xi_{ijk}\geq 0$  which absorbs small violations of the constraint~(\ref{eq:largemargin}). The resulting optimization problem can be formulated as the following semi-definite program (SDP)~\citep{BoyVan04}:
\begin{center}
\fbox{\parbox{0.50\textwidth}{
\vspace{1ex}
\hspace{1ex}\begin{tabular}{l}	
	$ \displaystyle \min_{\Mb\succeq 0} \displaystyle \sum_{j\leadsto i} d^2_{\Mb}(\x_i,\x_j) + \mu\!\! \displaystyle \sum_{j\leadsto i,k:y_k\neq y_i} \!\!\xi_{ijk} $\\
	\textbf{subject to}: \\
	\hspace{2ex}$(1) \hspace{4pt} d^2_{\Mb}(\x_i,\x_k) - d^2_{\Mb}(\x_i,\x_j) \ge 1 - \xi_{ijk}$\\
	\hspace{2ex}$(2) \hspace{4pt} \xi_{ijk} \ge 0$ \\
\end{tabular}
\vspace{1ex}
}}
\end{center}
Here $\mu\geq 0$ defines the trade-off between minimizing the objective and penalizing constraint violations (by default we set $\mu=1$). 

\subsection{Information-Theoretic Metric Learning}
Different from NCA and LMNN, Information-Theoretic Metric Learning (ITML), proposed by~\cite{Davis07itml}, does not minimize the leave-one-out error of kNN classification. In contrast, ITML assumes a uni-modal data-distribution and clusters similarly labeled inputs close together while regularizing the learned metric to be close to some pre-defined initial metric in terms of Gaussian cross entropy (for details see~\cite{Davis07itml}). 
Similar to LMNN, ITML also incorporates the similarity and dissimilarity as constraints in its optimization. Specifically, ITML enforces that similarly labeled inputs must have a distance smaller than a given upper bound $d_{\Mb}(\x_i,\x_j) \le u$ and dissimilarly labeled points must be further apart than a pre-defined lower bound $d_{\Mb}(\x_i,\x_j) \ge l$. 
If we denote the set of similarly labeled input pairs as $S$, and  dissimilar pairs as $D$, the optimization problem of ITML is:
\begin{center}
\fbox{
\hspace{1ex}\parbox{0.45\textwidth}{
\vspace{1ex}
\begin{tabular}{l}
	$\displaystyle \min_{\Mb\succeq 0}  \textrm{tr}(\Mb\Mb_0^{-1})-\log\det(\Mb\Mb_0^{-1})$  \\
	\textbf{subject to:}\\
	 \hspace{2ex} $(1)\  d^2_{\Mb}(\x_i,\x_j) \le u  \hspace{2ex} \forall (i,j)\in S,$ \\
	 \hspace{2ex} $(2)\  d^2_{\Mb}(\x_i,\x_j) \ge l \hspace{2.5ex} \forall (i,j)\in D.$\\
\end{tabular}
\vspace{1ex}
}}
\end{center}
\cite{Davis07itml} introduce several variations, including the incorporation of slack-variables. One advantage of the particular formulation of the  ITML optimization problem is that the SDP constraint $\Mb\succeq 0$ does not have to be monitored explicitly through eigenvector decompositions but is enforced implicitly through the objective.

\begin{table*}[ht]
    \tabcolsep 1.8pt
    \small
	\center	
\center
\begin{tabular}{|l|c|c|c|c|c|c|c|c|c|c|}
\hline {\bf Statistics} & {\bf Haber}  & {\bf Credit} & {\bf ACredit} & {\bf Trans} & {\bf Diabts} & {\bf Mammo} & {\bf CMC} & {\bf Page} & {\bf Gamma} \\
\hline
\hline \#examples & 306 &  653 & 690 & 748 & 768 & 830 & 962 & 5743 & 19020 \\
\hline \#features &  3 & 15 & 14 & 4 & 8 & 5 & 9 & 10 & 11\\
\hline \#training exam. &   245 & 522 & 552 &  599 & 614 & 664 & 770 & 4594 & 15216\\
\hline \#testing  exam. & 61 & 131 & 138 & 150 & 154 & 166 & 192  & 1149 & 3804  \\
\hline
\hline {\bf Metric}  & \multicolumn{9}{|c|}{\bf Error Rates} \\ 
\hline Euclidean   & 27.37 & {\bf 13.12} & {\bf 14.11} & {\bf 20.54} & 23.46 & 18.17 & 26.91 & {\bf 2.56} & {\bf 12.62} \\
\hline ITML & {\bf 26.50} & 13.68 & 14.71 & 22.86 & 23.14 & 18.20 & 27.67 & 4.78 & 21.50 \\
\hline NCA  & {\bf 26.39} & 13.48 & {\bf 14.10} & 22.59 & {\bf 22.74} & 18.17 & {\bf 26.53} & 4.74 & N/A \\
\hline LMNN   & {\bf 26.70} & 13.48 & {\bf 13.89} & 20.81 & {\bf 22.89} & {\bf 17.78} & {\bf 26.68} & {\bf 2.66} & 13.04 \\
\hline
\end{tabular}
\caption{Error rates of SVM classification with an RBF kernel (all parameters were set by 5-fold cross validation) under various learned metrics.\label{results:metric}}
\end{table*}

\subsection{Metric Learning for SVM}

\label{sec:results_metric}
We evaluate the efficacy of NCA, ITML and LMNN as pre-processing step for SVM classification with an RBF kernel. We used nine data sets from the UCI Machine Learning repository~\citep{Frank+Asuncion:2010} of varying size, dimensionality and task description. The data sets are: Haberman's Survival (Haber), Credit Approval (Credit), Australian Credit Approval (ACredit), Blood Transfusion Service (Trans), Diabetes (Diabts), Mammographic Mass (Mammo), Contraceptive Method Choice (CMC), Page Blocks Classification (Page) and MAGIC Gamma Telescope (Gamma). 

For simplicity, we restrict our evaluation to the binary case and convert multi-class problems to binary ones, either by selecting the two most-difficult classes or (if those are not known) by grouping labels into two sets. Table~\ref{results:metric} details statistic and classification results on all nine data sets. The best values up to statistical significance (within a 5\% confidence interval)  are highlighted in bold. To be fair to all algorithms, we re-scale all features to have standard deviation $1$. 
We follow the commonly used heuristic for Euclidean RBF\footnote{The choice of $\sigma^2=\#features$ is also the default value for the LibSVM toolbox~\citep{CC01a}.} and initialize NCA and ITML with $\Lb_0=\frac{1}{d}\Ib$ for all experiments (where $d$ denotes the $\#features$).  As LMNN is known to be very parameter insensitive, we set $\mu$ to the default value of $\mu=1$. All SVM parameters ($C$ and $\sigma^2$) were set by 5-fold cross validation on the training sets, after the metric is learned. The results on the smaller data sets $(n<1000)$ were averaged over 200 runs with random train/test splits, Page Blocks (Page) was averaged over 20 runs and Gamma was run once (here the train/test splits are pre-defined). 

In terms of scalability, NCA is by far the slowest algorithm and our implementation did not scale up to the (largest) Gamma data set. LMNN and ITML require comparable computation time (on the order of several minutes for the small- and 1-2 hours for large data sets -- for details see Section~\ref{sec_related}). As a general trend, none of the three metric learning algorithms consistently outperforms the Euclidean distance. Given the additional computation time, it is questionable if either one is a reasonable pre-processing step of SVM-RBF classification. This is in large contrast with the drastic improvements that these metric learning algorithms obtain when used as pre-processing for kNN~\citep{Goldberger05,Weinberger06distancemetric,Davis07itml}. One explanation for this discrepancy could be based on the subtle but important differences between the kNN classification rule~(\ref{eq:knn_rule}) and the one of SVMs~(\ref{eq:svm_h}). In the remainder of this paper we will explore the possibility to learn a metric explicitly for the SVM decision rule.

\section{Support Vector Metric Learning}
\label{sec_method}

As a first step towards learning a metric specifically for SVM classification, we incorporate the squared Mahalanobis distance~(\ref{eq:mahala})  into the kernel function (\ref{eq:rbf}) and define the resulting kernel function and matrix as
\begin{equation}
k_{\Lb}(\x_i,\x_j) = e^{-(\x_i-\x_j)^\top\Lb^\top\Lb(\x_i-\x_j)} \ \textrm{and} \ \Kb_{ij}=k_{\Lb}(\x_i,\x_j).\label{eq:kb}
\end{equation}
As mentioned before, the typical Euclidean RBF setting is a special case where $\Lb=\frac{1}{\sigma}\Ib^{d\times d}$. 


\subsection{Loss function}
\begin{wrapfigure}{r}{0.45\textwidth}
\vspace{-5ex}
	\centerline{\includegraphics[width=0.45\textwidth]{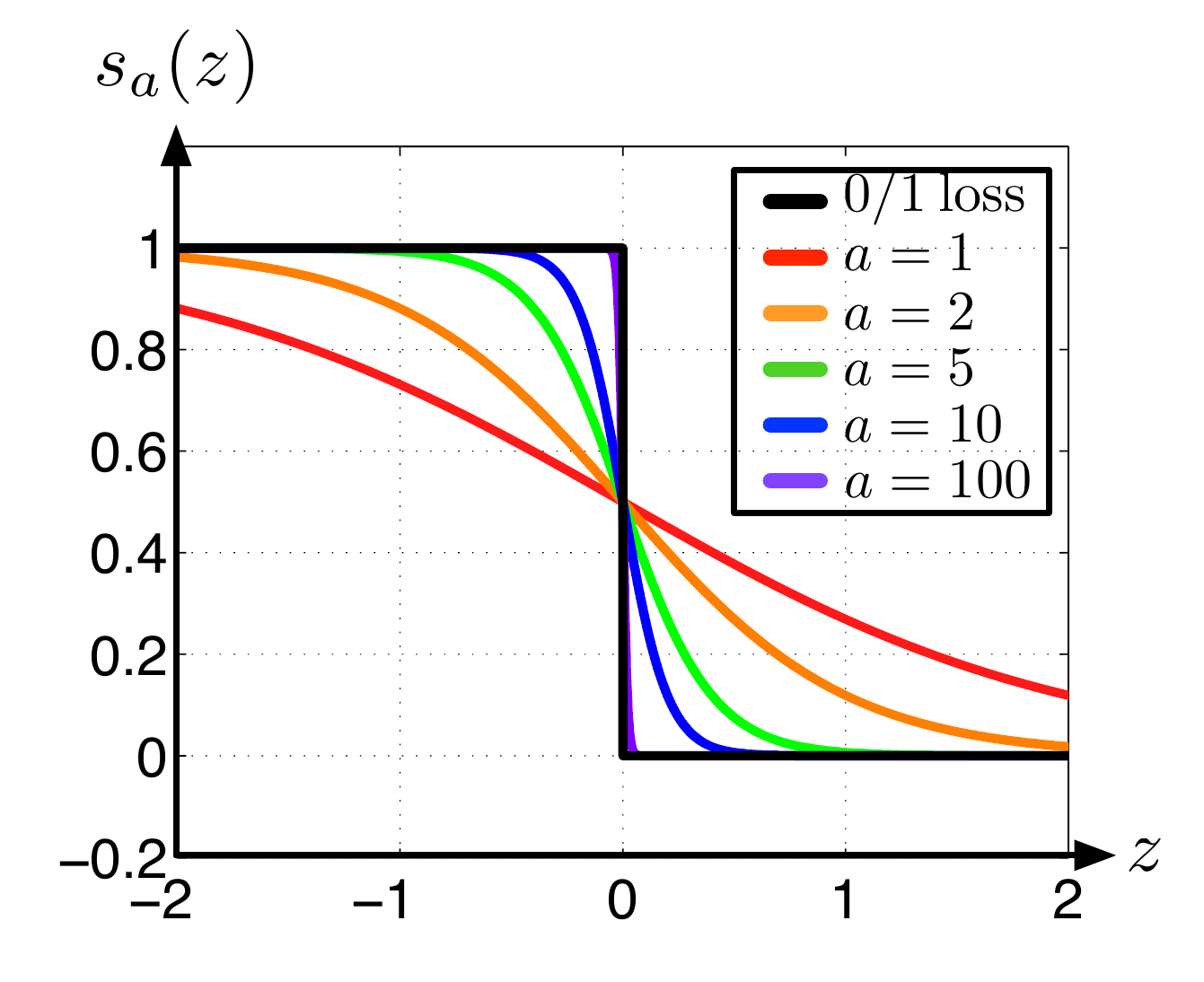}}
 	\caption{The function $s_a(z)$ is a soft (differentiable) approximation of the zero-one loss. The parameter $a$ adjusts the steepness of the curve.}
 	\label{fig:softloss}
	\vspace{-5ex}
\end{wrapfigure}
In the Euclidean case, a standard way to select the meta parameter $\sigma$ is through cross-validation. In its simplest form, this involves splitting the training data set into two mutually exclusive subsets: training set $T$ and validation set $V$. The SVM parameters  $\alpha_i,b$ are then trained on $T$ and the outcome is evaluated on the validation data set $V$. After a gridsearch over several candidate values for $\sigma$ (and $C$), the setting that performs the best on the validation data is chosen. For a single meta parameter, search by cross validation is simple and surprisingly effective. If more meta parameters need to be set --- in the case of choosing a matrix $\Lb$, this involves $d\times d$ entries --- the number of possible configurations grows exponentially and the gridsearch becomes infeasible.

We follow the intuition of validating meta parameters on a hold-out set of the training data. 
Ideally, we want to find a metric parameterized by $\Lb$ that minimize the classification error ${\cal{E}}_V$ on the validation data set
\begin{eqnarray}
	\Lb=\textrm{arg}\!\min_{\Lb}{\cal E}_V(\Lb) \label{eq:valloss} 
	\textrm{ where: }\ \nonumber 
 	{\cal E}_V(\Lb)=\frac{1}{|V|}\sum_{(\x,y)\in V} [h(\x)=y].
\end{eqnarray}
Here $[h(\x)=y]\in\{0,1\}$ takes on value $1$ if and only if $h(\x)=y$. 
The classifier $h(\cdot)$, defined in~(\ref{eq:svm_h}) depends on parameters $\alpha_i$ and $b$, which are re-trained for every intermediate setting of $\Lb$. Performing the minimization in~(\ref{eq:valloss}) is non-trivial because the sign$(\cdot)$ function in~(\ref{eq:svm_h}) is non-continuous. We therefore introduce a smooth loss function ${\cal L}_V$, which mimics ${\cal E}_V$, but is better behaved. 
\begin{equation}
	{\cal L}_V(\Lb)=\frac{1}{|V|}\sum_{(\x,y)\in V} s_a(yh(\x))\ 
	\textrm{ where: }\ s_a(z)=\frac{1}{1+e^{az}}. 
\end{equation}
The function $s_a(z)$ is the mirrored sigmoid function, a soft approximation of the zero-one loss. The parameter $a$ adjusts the steepness of the curve. In the limit, as $a\gg 0$ the function ${\cal L}_V$ becomes identical to ${\cal E}_V$. Figure~\ref{fig:softloss} illustrates the function $s_a(\cdot)$ for various values of $a$. 

\subsection{Gradient Computation}
Our surrogate loss function ${\cal L}_V$ is continuous and differentiable so we can compute the derivative $\frac{\partial{\cal L}_V}{\partial h(\x)}$. To obtain the derivative of ${\cal L}_V$ with respect to $\Lb$ we need to complete the chain-rule and also compute $\frac{\partial h(\x)}{\partial \Lb}$. The SVM prediction function $h(\x)$, defined in (\ref{eq:svm_h}), depends on $\Lb$ indirectly through $\mathbf{\alpha}_i,b$ and $\Kb$. In the next paragraph we follow the original approach of \citep{Chapelle02choosingmultiple} for kernel parameter learning. This approach has also been used successfully for wrapper-based multiple-kernel-learning~\citep{rakotomamonjy2008simplemkl,sonnenburg2006large,kloft2010non}. 
 For ease of notation,  we abbreviate $h(\x)$ by $h$ and use the vector notation $\alpha=[\alpha_1,\dots,\alpha_n]^\top$. 
%
Applying the chain-rule to the derivative of $h$ results in:
\begin{equation}
\frac{\partial {h}}{\partial \Lb} = \frac{\partial {h}}{\partial \alpha} 
				\frac{\partial \alpha}{\partial \Lb}  + \frac{\partial {h}}{\partial \Kb}  \frac{\partial \Kb}{\partial \Lb}  +
				\frac{\partial {h}}{\partial b} \frac{\partial b}{\partial \Lb}. 
\end{equation}

The derivatives $\frac{\partial {h}}{\partial \alpha },\frac{\partial {h}}{\partial b},\frac{\partial {h}}{\partial \Kb },\frac{\partial \Kb}{\partial\Lb}$ are straight-forward and follow from definitions~(\ref{eq:kb}) and (\ref{eq:svm_h})~\citep{peterson2008}.
In order to compute $\frac{\partial {\alpha}}{\partial \Lb}$ and $\frac{\partial {b}}{\partial \Lb}$ , we express the vector  $(\alpha,b)$  in closed-form with respect to $\Lb$. Because we absorb slack variables through our kernel modification in~(\ref{eq:ridge}) and we use a hard-margin SVM with the modified kernel, all support vectors must lie exactly one unit from the hyperplane and satisfy 
\begin{equation}
	y_i(\sum_{j=1}^n \Kb_{ij}\alpha_jy_j + b) = 1.\label{svm:sv}
\end{equation}
Since the parameters $\alpha_j$ of non-support vectors are zero, the derivative of these $\alpha_j$ with respect to ${\Lb}$ are also all-zero and do not need to be factored into our calculation. We can therefore (with a slight abuse of notation) remove all rows and columns of $\Kb$ that do not correspond to support vectors and express ~(\ref{svm:sv}) as a matrix equality
	\[ 
	\underbrace {
	\left( \begin{array}{ccc} 
	{\bar \Kb} & {\bf y} \\
	{\bf y^\top} & 0 \\
	\end{array} \right) }_{\Hb}
	\left( \begin{array}{ccc} 
	{\bf \alpha}\\
	b\\
	\end{array} \right)
	=
	\left( \begin{array}{ccc} 
	{\bf 1}\\
	0\\
	\end{array} \right)\]
where $\bar\Kb_{ij} = y_i y_j {\Kb}(x_i,x_j)$. Consequently, we can solve for $\alpha$ and $b$ through left-multiplication with  $\Hb^{-1}$. Further, the derivative with respect to $\Lb$ can be derived from the matrix inverse rule~\citep{peterson2008}, leading to
\begin{eqnarray}
({\bf \alpha},b)^\top \!=\! \Hb^{-1} (1\cdots1,0)^\top \ 
\textrm{ and }\ \frac{\partial (\alpha, b)}{\partial \Lb_{ij}} \!=\! -\Hb^{-1}\frac{\partial \Hb}{\partial \Lb_{ij}}(\alpha,b)^\top. \label{eq:abder}
\end{eqnarray} 

\begin{figure*}[t]
	\centering
		\includegraphics[width=5.5in]{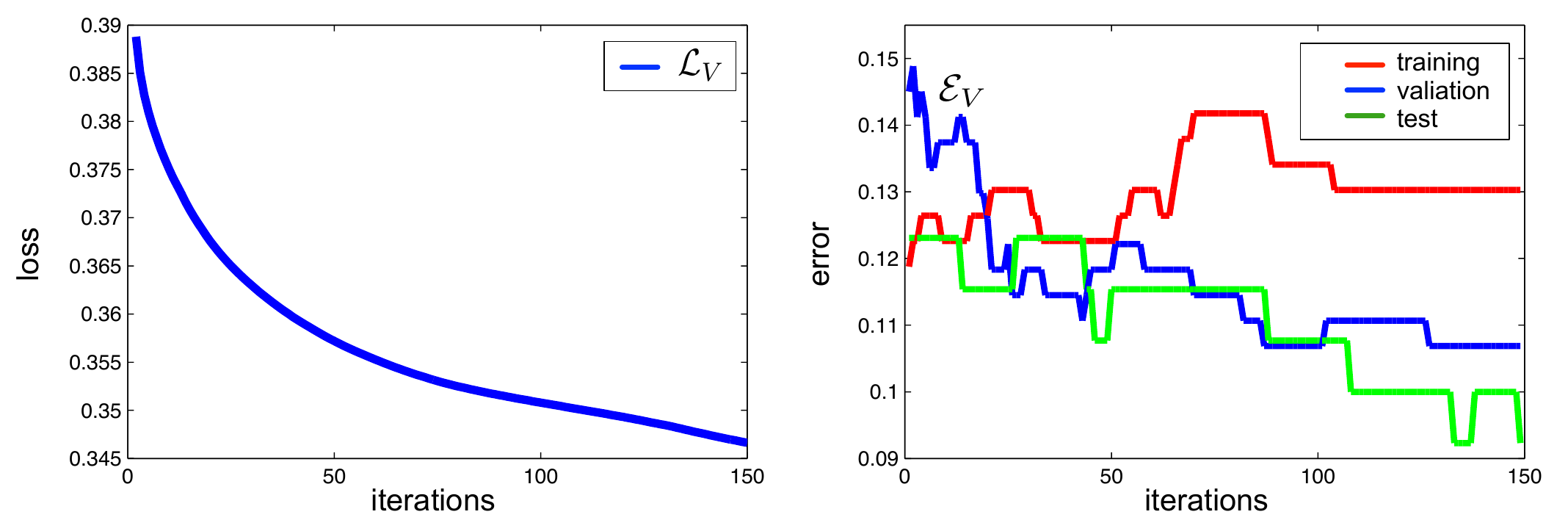}
	\caption{An example of training, validation and test error on the Credit data set. As the loss ${\cal L}_V$ (left) decreases, the validation error ${\cal E}_V$ (right) follows suit (solid blue lines). For visualization purposes, we did not use a second-order function minimizer but simple gradient descent with a small step-size.}
	\label{fig:trainvaltest}
\end{figure*}

\subsection{Optimization}
Because the derivative $\frac{\partial \Hb}{\partial \Lb}$ follows directly from the definition of $\bar\Kb$ and~(\ref{eq:kb}), this completes the gradient $\frac{\partial{\cal L}_V}{\partial \Lb}$. We can now use standard gradient descent, or second order methods to minimize (\ref{eq:valloss}) up to a local minimum. 
It is important to point out that~(\ref{eq:abder}) requires the computation of the optimal $\alpha,b$, given the current matrix $\Lb$. These can be obtained with any one of the many freely available SVM packages~\citep{CC01a} by solving the SVM optimization~(\ref{eq:svm_hard}) for the kernel $\Kb$ that results from $\Lb$.  
In addition, we also learn the regularization constant $C$ from eq.~(\ref{eq:ridge}) with our gradient descent optimization.  For brevity we omit the exact derivation of $\frac{\partial {\cal L}_V}{\partial C}$ but point out that it is very similar to the gradient with respect to $\Lb$, except that it is computed only from the diagonal entries of $\Kb$. 

We control the steps of gradient descent by early-stopping. We use part of the training data as a small hold-out set to monitor the algorithm's performance, and we stop the gradient descent when the validation results cease to improve. 

We refer to our algorithm as \emph{Support Vector Metric Learning} (SVML). Algorithm~\ref{table:algo} summarizes SVML in pseudo-code. Figure~\ref{fig:trainvaltest} illustrates the value of the loss function ${\cal L}_V$ as well as the training, validation and test errors.  

\begin{algorithm}[h!!!!]
\caption{SVML in pseudo-code.\label{table:algo}}
\label{algo}
\begin{algorithmic}[1]
\STATE Initialize ${\Lb}$.
\WHILE{Hold-out set result keeps improving}
\STATE Compute kernel matrix $\Kb$ from $\Lb$ as in (\ref{eq:mahala}).
\STATE Call SVM with $\Kb$ to obtain $\alpha$ and $b$. 
\STATE Compute gradient $\frac{\partial {\cal L}_V}{\partial\Lb}$ as in (\ref{eq:abder}) and perform update on $\Lb$.
\ENDWHILE
\end{algorithmic}
\end{algorithm}


\subsection{Regularization and Variations}
In total, we learn $d\times d$ parameters for the matrix $\Lb$ and $n+1$ parameters for $\alpha$ and $b$. 
To avoid overfitting, we add a regularization term to the loss function, which restricts the matrix ${\Lb}$ from deviating too much from its initial estimate ${\Lb}_{0}$:
\begin{eqnarray}
\label{eq:regul}
{\cal L}_V(\Lb)=\frac{1}{|V|}\sum_{(\x,y)\in V} s_a(yh(\x))+\lambda \| \Lb - \Lb_{0} \|_F^2
\end{eqnarray}

Another way to avoid overfitting is to impose structural restrictions on the matrix $\Lb$. If $\Lb$ is restricted to be spherical, $\Lb=\frac{1}{\sigma}\Ib^{d\times d}$, SVML reduces to kernel width estimation. Alternatively, one can restrict $\Lb$ to be any diagonal matrix, essentially performing feature re-weighing. This can also be useful as a method for feature selection in settings with noisy features~\citep{weston2001feature}. We refer to these two settings as SVML-Sphere and SVML-Diag. Both of these special scenarios have been studied in previous work in the context of kernel parameter estimation~\citep{ayat2005automatic,Chapelle02choosingmultiple}. See section~\ref{sec_related} for a discussion on related work. 

Another interesting structural limitation is to enforce $\Lb\in{\cal R}^{r\times d}$ to be rectangular, by setting $r<d$. This can be particularly useful for data visualization. For high dimensional data, the decision boundary of support machines is often hard to conceptualize. By setting $r=2$ or $r=3$, the data is mapped into a low dimensional space and can easily be plotted.


\subsection{Implementation} 
The gradient, as described in this section, can be computed very efficiently. We use a simple $C/Mex$ implementation with Matlab. As our SVM solver, we use the open-source Newton-Raphson implementation from Olivier Chapelle\footnote{Available at http://olivier.chapelle.cc/primal/.}. As function minimizer we use an  open-source implementation of conjugate gradient descent\footnote{Courtesy of Carl Edward Rasmussen, available from \url{http://www.gatsby.ucl.ac.uk/~edward/code/minimize/minimize.m}}. Profiling of our code reveals that over 95\% of the gradient computation time was spent calling the SVM solver. For a large-scale implementation, one could use special purpose SVM solvers that are optimized for speed~\citep{bottou2007large,joachims1999making}. Also, the only computationally intensive parts of the gradient outside of the SVM calls are all trivially parallelizable and could be computed on multiple cores or graphics cards. However, as it is besides the point of this paper, we do not focus on further scalability. 

\section{Results}
\label{sec_results}
To evaluate SVML, we revisit the nine data sets from Section~\ref{sec:results_metric}. For convenience, Table~\ref{results:small} restates all relevant data statistics and also includes classification accuracies for all metric learning algorithms. SVML is naturally slower than SVM with Euclidean distance but requires no cross validation for any meta parameters. For better comparison, we also include results for 1-fold and 5-fold cross validation for all other algorithms.  In both cases, the meta parameters $\sigma^2,C$ were selected from five candidates each -- resulting in 25 or 125 SVM executions. The kernel width $\sigma^2$ is selected from within the set $\{4d, 2d,d,\frac{d}{2}, \frac{d}{4}\}$ and the meta parameter $C$ was chosen from within $\{0.1,1,10,100\}$. 
As SVML is not particularly sensitive to the exact choice of $\lambda$ -- the regularization parameter in (\ref{eq:regul}) -- we set it to $100$ for the smaller data sets ($n<1000$) and to $10$ for the larger ones  (\texttt{Page}, \texttt{Gamma}). We terminate our algorithm based on a small hold-out set.



\begin{table*}[ht]
    \tabcolsep 1.8pt
    \small
	\center	
\center
\begin{tabular}{|l|c|c|c|c|c|c|c|c|c|c|}
\hline {\bf Statistics} & {\bf Haber}  & {\bf Credit} & {\bf ACredit} & {\bf Trans} & {\bf Diabts} & {\bf Mammo} & {\bf CMC} & {\bf Page} & {\bf Gamma} \\
\hline
\hline \#examples & 306 &  653 & 690 & 748 & 768 & 830 & 962 & 5743 & 19020 \\
\hline \#features &  3 & 15 & 14 & 4 & 8 & 5 & 9 & 10 & 11\\
\hline \#training exam. &   245 & 522 & 552 &  599 & 614 & 664 & 770 & 4594 & 15216\\
\hline \#testing  exam. & 61 & 131 & 138 & 150 & 154 & 166 & 192  & 1149 & 3804  \\
\hline
\hline {\bf Metric}  & \multicolumn{9}{|c|}{\bf Error Rates} \\ 
\hline Euclidean 1-fold & 27.16 & 13.16 & 14.36 & 21.05 & 23.84 & 18.43 & 27.12 & {\bf 2.61} & 12.70 \\
\hline Euclidean 3-fold & 27.40 & 13.10 & 14.13 & 20.58 & 23.39 & 18.27 & 26.77 & {\bf 2.55} & 12.68 \\
\hline Euclidean 5-fold & 27.37 & 13.12 & 14.11 & 20.54 & 23.46 & 18.17 & 26.91 & {\bf 2.56} & {\bf 12.62} \\
\hline ITML + SVM 1-fold & 26.57 & 13.78 & 14.15 & 23.01 & 23.19 & 19.14 & 28.65 & 4.82 & 22.63 \\
\hline ITML + SVM 3-fold & {\bf 26.13} & 13.58 & {\bf 13.88} & 22.98 & 23.17 & 17.98 & 27.68 & 4.77 & 21.50 \\
\hline ITML + SVM 5-fold & 26.50 & 13.68 & 14.71 & 22.86 & 23.14 & 18.20 & 27.67 & 4.78 & 21.50 \\
\hline NCA + SVM 1-fold & 26.44 & 13.74 & 14.14 & 22.89 & {\bf 22.84} & 17.76 & 27.47 & 4.73  & N/A\\
\hline NCA + SVM 3-fold & 26.47 & 13.45 & 14.00 & 22.67 & {\bf 22.72} & 18.12 & 26.60 & 4.73  & N/A\\
\hline NCA + SVM 5-fold & 26.39 & 13.48 & 14.10 & 22.59 & {\bf 22.74} & 18.17 & {\bf 26.53} & 4.74 & N/A \\
\hline LMNN + SVM 1-fold & {\bf 26.38} & 13.11 & 13.97 & 21.02 & 22.97 & 17.84 & 26.80 & 2.85 & 13.04 \\
\hline LMNN + SVM 3-fold & 26.44 & 13.30 & {\bf 13.93} & 20.73 & {\bf 22.86} & {\bf 17.57} & 26.66 & 2.81 & 12.79 \\
\hline LMNN + SVM 5-fold & 26.70 & 13.48 & {\bf 13.89} & 20.81 & {\bf 22.89} & 17.78 & 26.68 & {\bf 2.66} & 13.04 \\
\hline SVML-Sphere & 27.42 & 13.43 & {\bf 13.78} & {\bf 20.26} & 23.24 & 17.81 & 28.23 & 3.61 & 12.70 \\
\hline SVML-Diag & 28.15 & 13.33 & 15.11 & 20.46 & 24.14 & {\bf 17.35} & 29.51 & 2.92 & {\bf 12.54} \\
\hline SVML & {\bf 25.99} & {\bf 12.83} & {\bf 13.92} & 20.89 & 23.25 & {\bf 17.57} & {\bf 26.34} & 3.41 & {\bf 12.54} \\
\hline
\end{tabular}
\caption{Statistics and error rates for all data sets. The data sets are sorted by smallest to largest from left to right. The table shows statistics of data sets and error rates of SVML and comparison algorithms. The best results (up to a 5\% confidence interval) are highlighted in bold.\label{results:small}}
\end{table*}

As in Section~\ref{results:small}, experimental results are obtained by averaging over multiple runs on randomly generated 80/20 splits of each data set. For small data sets, we average 200 splits, 20 for medium size, and 1 for the large data set Gamma (where train/test splits are pre-defined). For the SVML training, we further apply a 50/50 split for training and validation within the training set, and another 50/50 split on the validation set for early stopping. The result from SVML appeared fairly insensitive to these splits.



As a general trend, SVML with a full matrix obtains the best results (up to significance) on $6$ out of the $9$ data sets. It is the only metric that consistently outperforms Euclidean distances. The diagonal version SVML-Diag and SVML-Sphere both obtain best results in $2$ out of $9$ and are not better than the uninformed Euclidean distance with 5-fold cross validation. None of the kNN metric learning algorithms perform comparably. 

\begin{figure}[t]
	\centering
		\includegraphics[width=\textwidth]{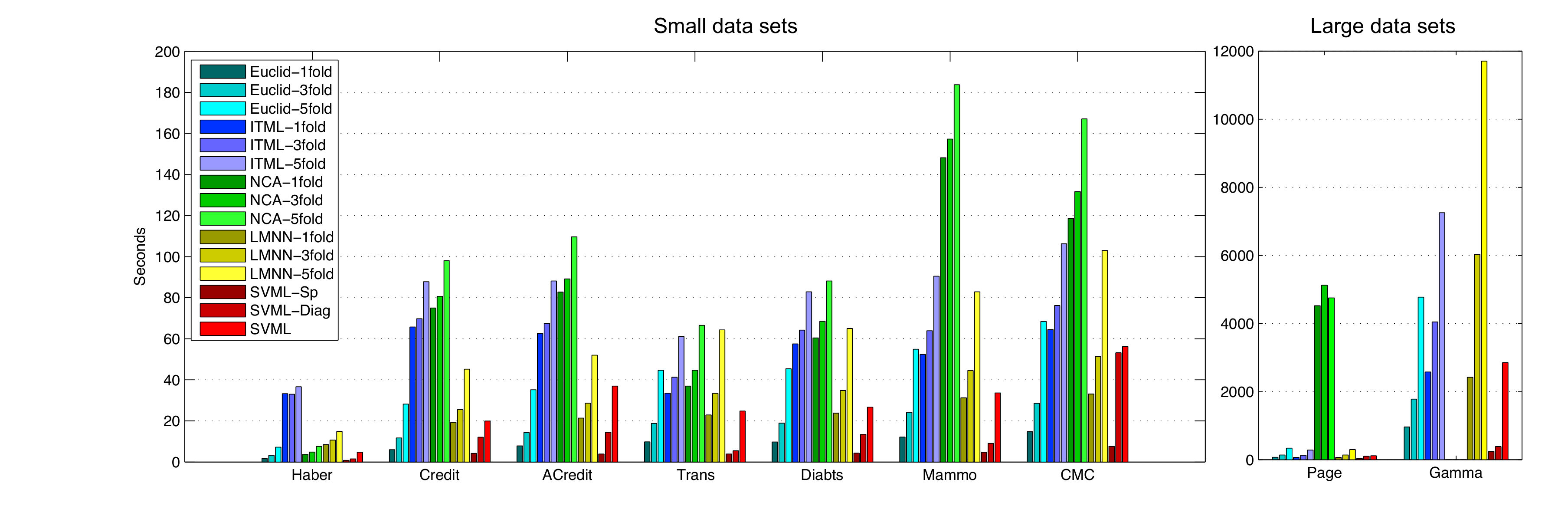}
	\caption{Timing results on all data sets. The timing includes metric learning, SVM training and cross validation. The computational resources for SVML training are roughly comparable with 3-5 fold cross validation with a Euclidean metric. (NCA did not scale to the Gamma data set.)  }
	\label{fig:time}
\end{figure}

In general, we found the time required for SVML training to be roughly between 3-fold and 5-fold cross validation for Euclidean metrics, usually outperforming LMNN, ITML and NCA. Figure~\ref{fig:time} provides running-time details on all data sets. We consider the small additional time required for SVML over Euclidean distances with cross validation as highly encouraging. 

\subsection{Dimensionality Reduction.} 
\begin{figure}[t]
	\centering
		\includegraphics[width=2.5in]{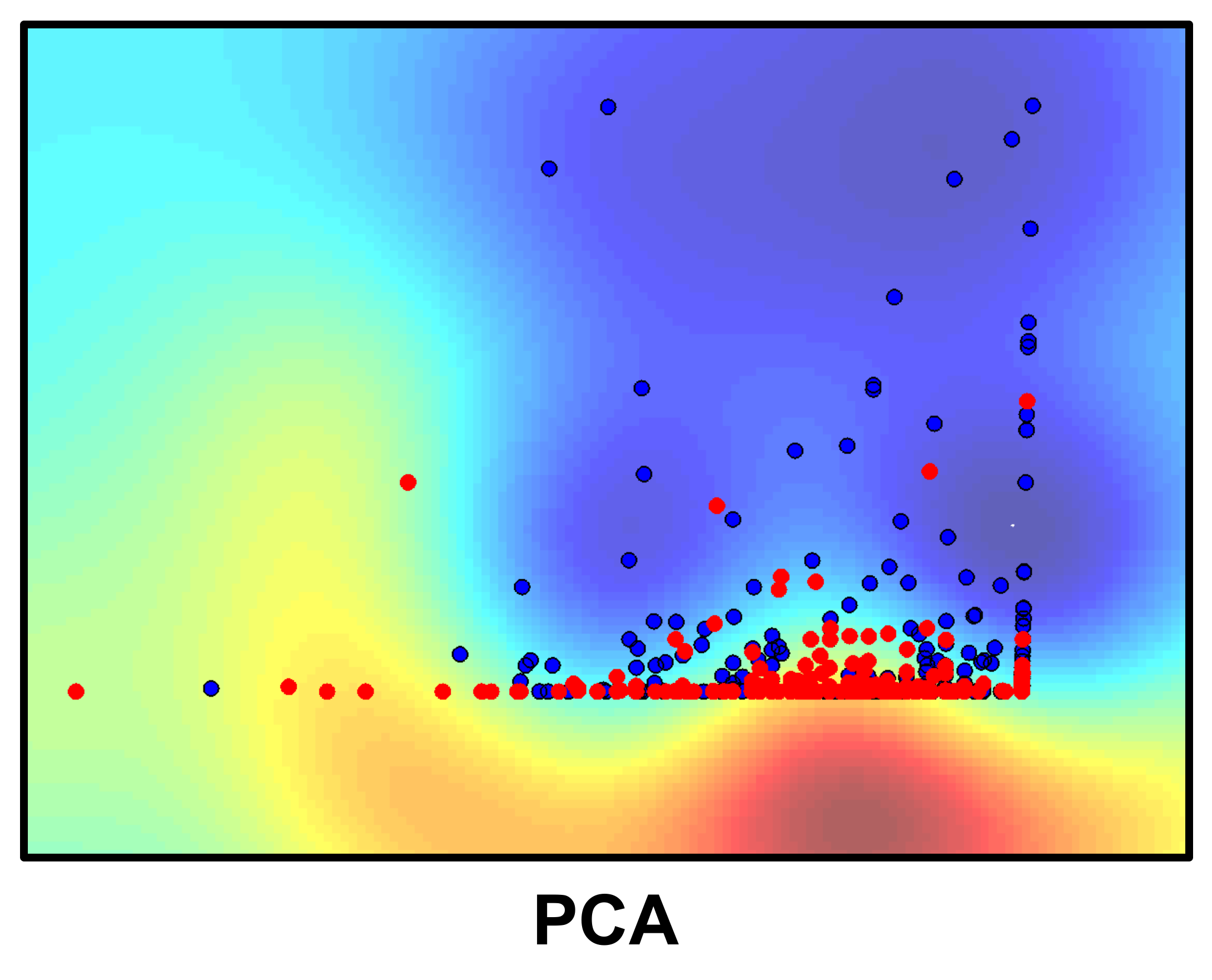}
		\includegraphics[width=2.5in]{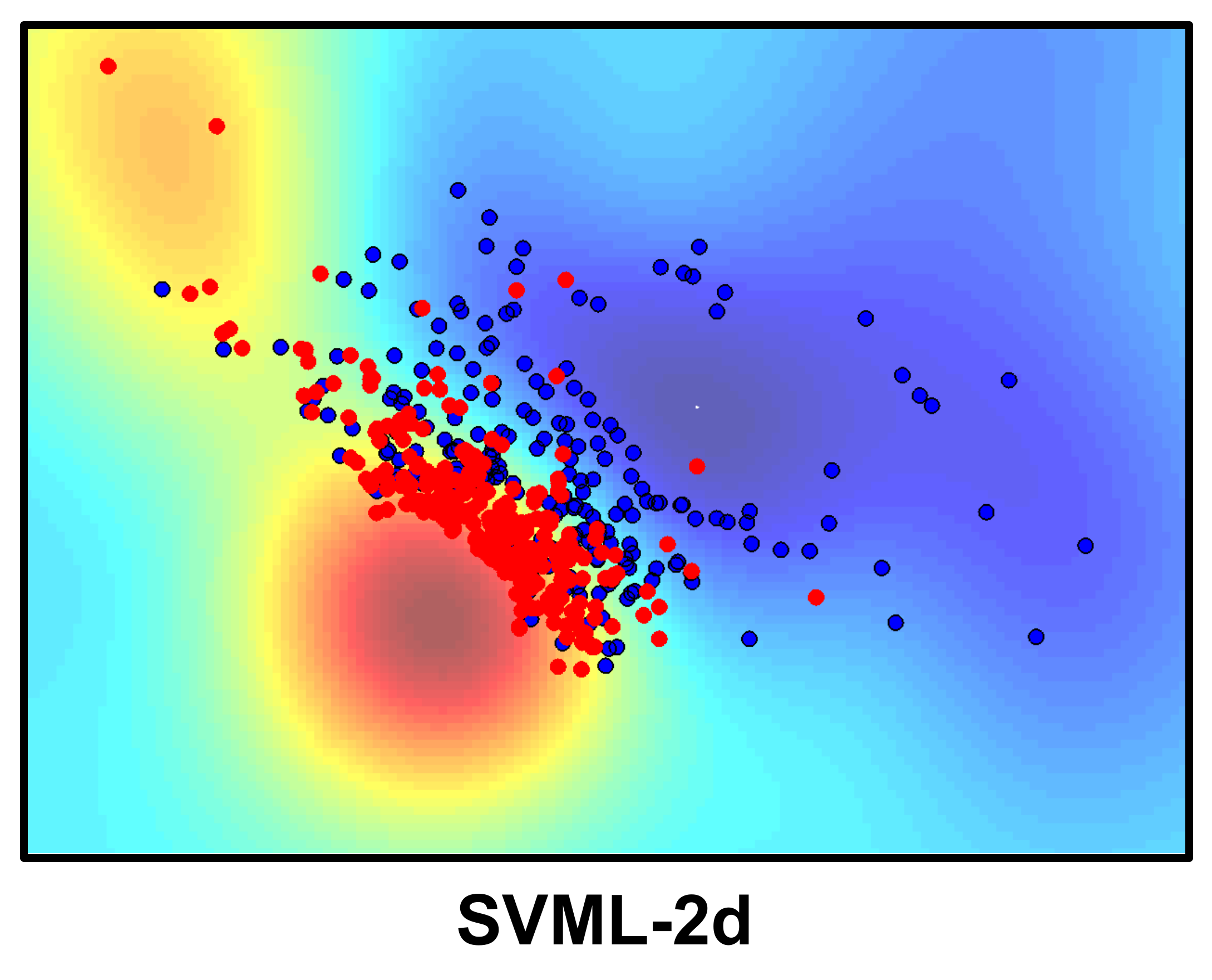}
	\caption{$2D$ visualization of the Credit data set. The figure shows the decision surface and support vectors generated by SVML ($\Lb\in{\cal R}^{2\times d}$) and standard SVM after projection onto the two leading principal components. }
	\label{fig:pca_svca}
\end{figure}

In addition to better classification results, SVML can also be used to map data into a low dimensional space while learning the SVM, allowing effective visualizations of SVM decision boundaries even for high dimensional data. To evaluate the capabilities of our algorithm for dimensionality reduction and visualization, we restrict $\Lb$ to be rectangular. Specifically, a mapping into a $r=2$ or $r=3$ dimensional space. 
As comparison, we use PCA to reduce the dimensionality before the SVM training without SVML (all meta parameters were set by cross-validation). Figure~\ref{fig:pca_svca} shows the visualization of the support vectors of the Credit data set after a mapping into a two dimensional space with SVML and PCA. The background is colored by the prediction function $h(\cdot)$.  
The $2D$ visualization shows a much more interpretable decision boundary. (Visualizations of the LMNN and NCA mappings were very similar to those of PCA.) Visualizing the support vectors and the decision boundaries of kernelized SVMs can help demystify hyperplanes in reproducing kernel Hilbert spaces and might help with data analysis.

\section{Related Work}
\label{sec_related}
Multiple publications introduce methods to learn Mahalanobis metrics. Previous work has focussed primarily on Mahalanobis metrics for $k$-nearest neighbor classifiers~\citep{Davis07itml,Globerson06,Goldberger05,Shental02,Shalev-Shwartz04,Weinberger06distancemetric} and clustering~\citep{Davis07itml,Shalev-Shwartz04,Shental02,Xing02}. None of these algorithms is specifically geared towards SVM classification. A detailed discussion of NCA, ITML and LMNN is provided in Section~\ref{sec_metric}. 


Another related line of work focusses on learning of the kernel matrix. The most common approach is to find convex combinations of already existing kernel matrices~\citep{bach2004multiple,lanckriet2004learning} or kernel learning through semi-definite programming~\citep{graepel2002kernel,ong2005learning}. 
The most similar area of related work is the field of kernel parameter estimation~\citep{ayat2005automatic,Chapelle02choosingmultiple,cherkassky2004practical,friedrichs2005evolutionary}. In particular, ~\citep{friedrichs2005evolutionary} can be viewed as learning a Mahalanobis metric for the Gaussian kernel --  however, instead of minimizing a soft surrogate of the validation error with gradient descent, the authors use genetic programming to maximize the ``fittness'' of the kernel parameters. The method of \citep{Chapelle02choosingmultiple} uses gradient descent to learn the $\sigma$ parameter of the RBF kernel matrix. SVML was highly inspired by this work. The main difference between our work and~\citep{Chapelle02choosingmultiple} is  that SVML learns the full matrix $\Lb$, and therefore a Mahalanobis metric, whereas Chapelle et al. only learn the parameter $\sigma$ or individual weights for blocks of features. Spherical and diagonal SVML can be viewed as a version of~\citep{Chapelle02choosingmultiple}. Similarly,~\citep{ayat2005automatic,schittkowski2005optimal} also explore feature re-weighting for support vector machines with alternative loss functions.

\section{Conclusion}
\label{sec_conclusion}
In this paper we investigate metric learning for SVMs. An empirical study of three of the most widely used out-of-the-box metric learning algorithms for kNN classification shows that these are not particularly well suited for SVMs. As an alternative, we derive SVML, an algorithm that seamlessly combines support vector classification with distance metric learning. SVML learns a metric that attempts to minimize the validation error of the SVM prediction at the same time as it trains the SVM classifier. On several standard benchmark datasets we demonstrate that our algorithm achieves state-of-the-art results with very high reliability. 
An important feature of SVML is that it is very insensitive to its few parameters (which we all set to default values) and does not require any model selection by cross validation. In fact, we demonstrate that SVML outperforms traditional SVM-RBF with the Euclidean distance (where parameters are set through cross validation)  consistently in accuracy while requiring a comparable amount of computation time. These aspects make SVML a very promising general-purpose metric learning algorithm for SVMs with RBF kernels, which also incorporates automatic model selection. 
We are currently implementing an open-source plug-in for the popular LIBSVM library~\citep{CC01a} and extending it to multi-class settings. 


\subsection{Acknowledgements}
We thank Marius Kloft, Ulrich Rueckert, Cheng Soon Ong, Alain Rakotomamonjy, Soeren Sonnenburg and Francis Bach for motivating this work. This material is based upon work supported by the National Institute of Health under grant NIH 1-U01-NS073457-01, and the National Science Foundation under Grant No. 1149882. Any opinions, findings, and conclusions or recommendations expressed in this material are those of the author(s) and do not necessarily reflect the views of the National Institute of Health and the National Science Foundation. Further, we would like to thank Yahoo Research for their generous support through the Yahoo Research Faculty Engagement Program. 

\small
\bibliographystyle{icml2011}
\bibliography{klearning}

\end{document}